\documentclass[conference]{IEEEtran}
\IEEEoverridecommandlockouts
\usepackage{cite}
\usepackage{amsmath,amssymb,amsfonts}
\usepackage{algorithmic}
\usepackage{graphicx}
\usepackage{textcomp}
\usepackage{caption}
\usepackage{subcaption}
\usepackage{xcolor}
\usepackage{balance}
\def\BibTeX{{\rm B\kern-.05em{\sc i\kern-.025em b}\kern-.08em
    T\kern-.1667em\lower.7ex\hbox{E}\kern-.125emX}}

\makeatletter

\let\old@ps@headings\ps@headings
\let\old@ps@IEEEtitlepagestyle\ps@IEEEtitlepagestyle
\def\confheader#1{%
    \def\ps@IEEEtitlepagestyle{%
        \old@ps@IEEEtitlepagestyle%
        \def\@oddhead{\strut\hfill#1\hfill\strut}%
        \def\@evenhead{\strut\hfill#1\hfill\strut}%
    }%
    \ps@headings%
}
\makeatother

\confheader{%
        \parbox{\textwidth}{\centering This article has been accepted for publication in the International Conference on Unmanned Aircraft Systems (ICUAS), 2025.}
}

\usepackage[pscoord]{eso-pic}
\newcommand{\placetextbox}[3]{
\setbox0=\hbox{#3}
\AddToShipoutPictureFG{ \put(\LenToUnit{#1\paperwidth},\LenToUnit{#2\paperheight}){\vtop{{\null}\makebox[0pt][c]{#3}}}
}
}
\placetextbox{.5}{0.060}{\footnotesize { \copyright Authors. Personal use of this material is permitted. Permission must be obtained for all other uses, in any current or future media } }
\placetextbox{.5}{0.050}{\footnotesize {including reprinting/republishing this material for advertising or promotional purposes, creating new  collective works, }}
\placetextbox{.5}{0.040}{\footnotesize {for resale or redistribution to servers or lists, or reuse of any copyrighted component of this work in other works.}}

\begin{document}


\title{\vspace{10mm}UAV Control with Vision-based Hand Gesture Recognition over Edge-Computing}

\author{\IEEEauthorblockN{
\large Sousannah Abdalla and Sabur Baidya\\
\normalsize Department of Computer Science and Engineering, University of Louisville, KY, USA}
\normalsize {e-mail:  sousannahmagdy@icloud.com,  sabur.baidya@louisville.edu}
\vspace{-4mm}
}

\maketitle

\begin{abstract}
Gesture recognition presents a promising avenue for interfacing with unmanned aerial vehicles (UAVs) due to its intuitive nature and potential for precise interaction. This research conducts a comprehensive comparative analysis of vision-based hand gesture detection methodologies tailored for UAV Control. The existing gesture recognition approaches involving cropping, zooming, and color-based segmentation, do not work well for this kind of applications in dynamic conditions and suffer in performance with increasing distance and environmental noises.
We propose to use a novel approach leveraging hand landmarks drawing and classification for gesture recognition based UAV control.
With experimental results we show that our proposed method outperforms the other existing methods
in terms of accuracy, noise resilience, and efficacy across varying distances, 
thus providing robust  control decisions.
However, implementing the deep learning based compute intensive gesture recognition algorithms on the UAV's onboard computer is significantly challenging in terms of performance. Hence, we propose to use a edge-computing based framework to offload the heavier computing tasks, thus achieving closed-loop real-time performance. With implementation over AirSim simulator as well as over a real-world UAV, we showcase the advantage of our end-to-end gesture recognition based UAV control system.  

\end{abstract}


\section{Introduction}

Unmanned Aerial Vehicles (UAVs) have revolutionized various industries, including surveillance, agriculture, and disaster management, due to their versatility and accessibility~\cite{muchiri2022review, daud2022applications}. One critical aspect of UAV operation is the interface between the operator and the vehicle. Traditional interfaces such as remote controllers or joysticks can be cumbersome and not intuitive, especially for non-expert users. Fully autonomous UAVs while can operate well with pre-planned mission, any dynamic changes in the flying environment may need human intervention for safety~\cite{elmokadem2021towards, stegagno2014semi}. Hence, gesture recognition based UAV control can be a useful alternative that can control the UAV in dynamic environment without using any radio controller, thus can work well even in presence of wireless interference when radio controllers can fail. Other applications of gesture-controlled UAVs include assistive applications where small UAVs can follow a human to provide information for assistance, and follow instructions. 

Gesture recognition allows users to control UAVs using natural hand movements. This approach not only simplifies the control mechanism but also enhances user experience and situational awareness. Although gesture-recognition based robot control has been used before using many wearable sensor technologies~\cite{patrona2021overview, yu2022end, yoo2022motion}, for high speed robots, e.g., UAVs which is used for various indoor and outdoor applications, the vision-based gesture recognition provides hassle-free control without the user needing any wearable sensors.
Vision-based gesture recognition, thus, holds significant potential for UAV control, leveraging computer vision algorithms to interpret hand gestures captured by onboard cameras or other visual sensors. However, there are several technical challenges involved to achieve this for this closed-loop UAV control as the UAV moves rapidly and distance from the object along with surrounding environments can change rapidly as well. So, we need to finetune a gesture recognition model that can be robust against these dynamics. 

In this paper, we present a comprehensive comparative analysis of four hand gesture detection methodologies tailored for UAV Control, focusing on vision-based gesture recognition. Alongside three methodologies previously examined in literature for hand gesture recognition, we introduce and evaluate a novel approach that emphasizes hand landmarks drawing for more accurate classification, thus results in more precise control decisions.
We evaluate the performance of each methodology based on criteria such as accuracy, noise resilience, and performance across varying distances, crucial for UAV operation. 
Through this analysis, we aim to provide insights into the strengths and limitations of each approach, facilitating informed decision-making for UAV control system design and implementation.

Implementing the proposed gesture recognition models on the UAVs are extremely challenging as the deep learning based algorithms are computationally intensive. To resolve this, we leverage an edge-assisted framework~\cite{jiang2019toward} where the heavier part of the computing can be offloaded to an edge server to ensure low-latency along with high accuracy in gesture recognition. This in turn reduces the end-to-end closed loop delay, thus, providing better control.
By advancing our understanding of hand gesture detection methods for UAV control along with edge-assisted distributed computing, this research contributes to the development of more intuitive and efficient interfaces, ultimately enhancing the usability and effectiveness of unmanned aerial systems across diverse applications. We have integrated this gesture control interface with AirSim UAV simulator~\cite{shah2018airsim} and also on a real-world drone.

The Summary of our contributions are as follows:

\begin{itemize}
    \item \textbf{Custom-trained hand gesture landmark model:} We developed a novel landmark-based hand gesture recognition model trained on a locally collected dataset of six classes, each with $1,500$ images, achieving a high accuracy rate of $96.14\%$. To the best of our knowledge landmark based hand gesture recognition has not been used before for detection with varying distances as in case of UAVs.

    \vspace{1mm}
    \item \textbf{Extended detection range using YOLOv4:} By integrating YOLOv4 with our landmark model, we successfully expanded the gesture detection range from $5$\,m to $10$\,m, enabling UAV operability in larger areas. We found no other existing vision-based hand gesture detection model can operate up to this distance.

    \vspace{1mm}
    \item \textbf{Latency reduction through edge computing:} Our edge computing framework allows significant reduction in processing delay, by leveraging offloading of the computation of the gesture recognition algorithm to an edge server, thus, maintaining latency below $30$\,ms, ensuring safe and responsive UAV control in real-time.
\end{itemize}

\vspace{3mm}
\section{Related Work}
In recent years, numerous approaches have emerged for UAV control systems, focusing on different control paradigms, including remote controllers, brain-computer interfaces, and gesture recognition.
For instance, Hu and Wang \cite{hu2020deep} developed a hand gesture recognition system utilizing deep learning techniques for UAV control. Their method converts dynamic gesture sequences into 2D matrices and 1D arrays, allowing the UAV to recognize and respond to specific gestures. While effective, the system’s accuracy can be affected by environmental factors such as lighting and background noise.
Samotyy et al. \cite{samotyy2024gesture} proposed a real-time gesture-based control system for quadcopters, using a deep learning model optimized for low computational requirements. Their approach enables efficient real-time recognition; however, its accuracy may drop in cluttered environments or when gestures are partially occluded.

Bello et al. \cite{bello2023captainglove} introduced CaptAinGlove, a glove-based gesture recognition system that uses capacitive and inertial sensors to control drones. The system ensures reliable recognition while maintaining low power consumption, but wearing an additional device may limit its usability in some applications.
Perera et al. \cite{perera2018uav} developed UAV-GESTURE, a dataset designed for UAV control using hand gestures in real-world outdoor settings. Their work provides a benchmark for evaluating gesture recognition models in drone applications, though achieving robust performance across different environmental conditions remains a challenge.
Lee and Yu \cite{lee2023wearable} proposed a multi-modal UAV control system combining hand gesture recognition with vibrotactile feedback. Their wearable controller uses inertial measurement units (IMUs) to classify hand gestures and provides real-time feedback via vibration motors, enhancing user experience and situational awareness.

Despite these advancements, challenges persist in developing gesture recognition methodologies that ensure high accuracy and robustness across various environmental conditions. Our work aims to bridge this gap by evaluating multiple gesture detection methods, with a focus on hand landmarks, to identify a more reliable solution for real-world UAV control scenarios.

\section{System Model and Challenges}

The proposed gesture recognition based UAV control framework is a closed-loop system as it involves sensing, processing, decision-making and control in real-time and it also feeds back the status of the UAV for the next iteration. Hence, there are several constraints and challenges involved in the system.

\begin{figure}[!t]
    \centering
    \includegraphics[width=0.6\linewidth]{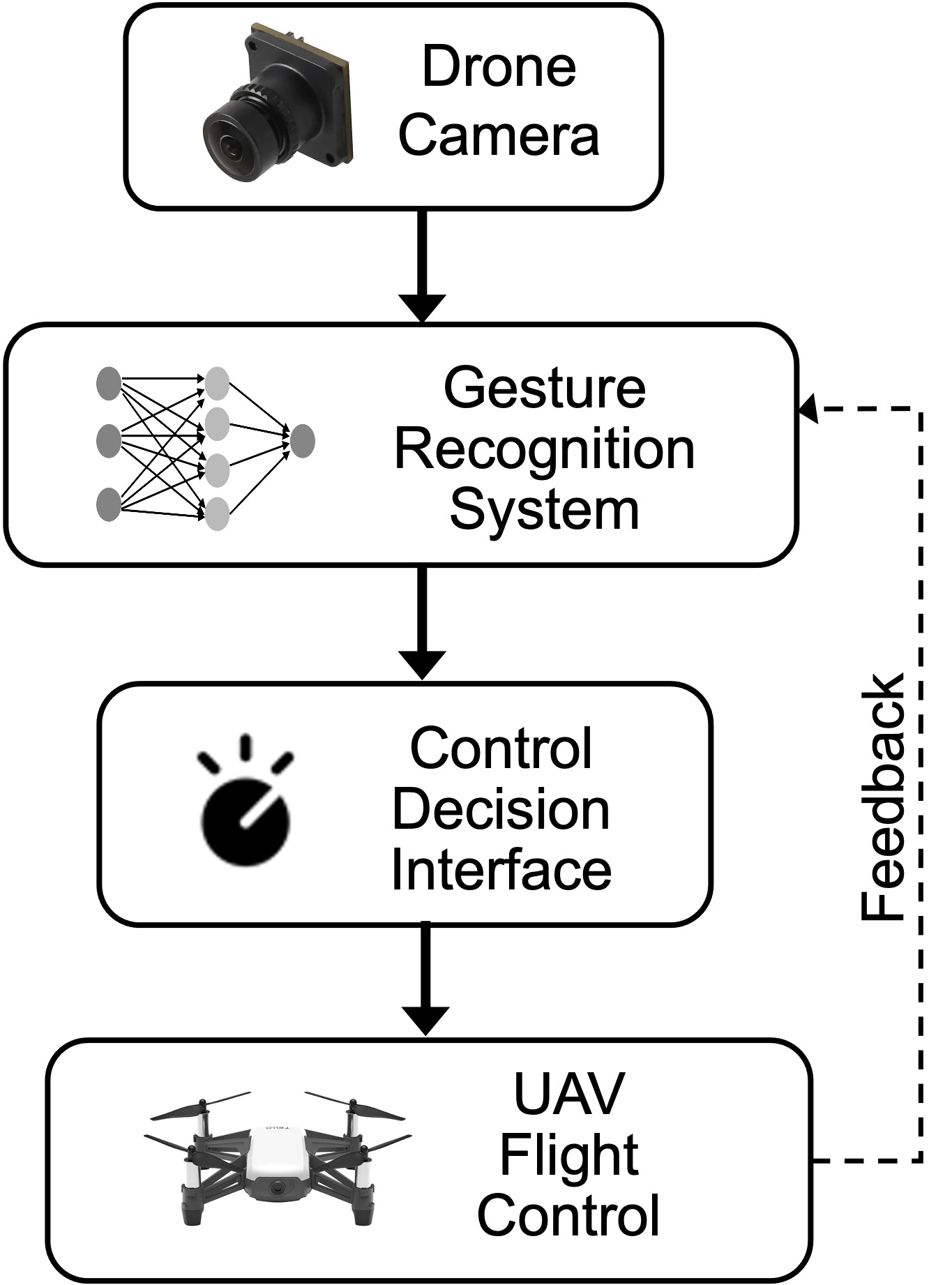}
    \vspace{2mm}
    \caption{System Model for Closed-Loop Gesture recognition based UAV Control}
    \label{fig:system_model}
\end{figure}

\subsection{System Model for Closed-Loop Gesture Control}

Figure \ref{fig:system_model} illustrates the closed-loop gesture control system designed to enable user interaction with a UAV through intuitive hand gestures. The system comprises three main components: the gesture recognition system, the control decision interface, and the UAV flight control.

In this model, the \textbf{gesture recognition system} utilizes computer vision techniques to capture hand gestures using the data captured by the front camera of the UAV. The camera continuously captures image frames, which are processed in real-time to identify specific gestures based on machine learning algorithms. 
Once a gesture is recognized, it is transmitted to the \textbf{control decision interface}, which translates the gesture data into control commands for the UAV. This interface serves as the bridge that converts recognized gestures into actionable instructions that the UAV can understand.

The \textbf{unmanned aerial vehicle (UAV) flight controller} receives these control commands and executes them to perform maneuvers based on the recognized gestures. The UAV operates under a closed-loop control system, continuously monitoring its position and performance to provide feedback to the user. This feedback mechanism allows for real-time adjustments, enhancing the user's control over the UAV.

To maintain effective communication and coordination among these components, we establish a decision time slot of $1$ second. During this interval, the UAV processes incoming commands, executes them, and sends back status updates, ensuring synchronized operation. The system is designed to minimize the deviation  $|\Delta x|$ between the desired and actual positions of the UAV, accounting for dynamic user interactions.

\subsection{Challenges in Gesture Recognition for UAV Control}

The implementation of a closed-loop gesture control system involves several challenges that must be addressed to ensure effective and reliable operation:

\begin{itemize}
    \item \textbf{Accuracy of Gesture Recognition}: Achieving high accuracy in gesture recognition is critical. Variations in hand shape, size, and the presence of obstructions can significantly impact the system's ability to recognize gestures correctly.

    \vspace{1mm}
    \item \textbf{Dynamic Distance Variation}: The distance between the user's hand and the UAV can change dynamically, affecting the performance of gesture recognition algorithms. The system must be robust enough to adapt to varying distances without compromising accuracy.

    \vspace{1mm}
    \item \textbf{Latency}: Low latency is crucial for maintaining a responsive control interface. Any delay in gesture recognition and command execution can lead to a poor user experience or even unsafe UAV operation. The system must minimize the latency to ensure real-time responsiveness.

    \vspace{1mm}
    \item \textbf{Environmental Factors}: External factors such as lighting conditions, background noise, and obstacles can interfere with the camera's ability to accurately capture and process hand gestures. The system must be designed to operate effectively under varying environmental conditions.

    \vspace{1mm}
    \item \textbf{User Training}: Users may require training to perform gestures consistently and accurately. The system must accommodate a learning curve and allow for adjustments in gesture definitions as needed.
\end{itemize}

By addressing these challenges, the closed-loop gesture control system can provide a reliable and intuitive method for UAV operation, enhancing user interaction and control.

\section{Methodology}
In this study, we focus on constructing a robust computer vision based gesture recognition algorithm for the UAV control. Different from other applications, that use the vision based gesture recognitions, e.g., human activity detection, our application has additional challenges due to the fast motion of the drone. If the UAV moves away from the human, the distance will increase rapidly along with the surrounding background, making it challenging for the computer vision algorithm to perform well. 
We explored various methods to develop an efficient hand gesture detection method that can achieve high accuracy even when the UAV moves away from the user up to a certain distance, and also have robustness against changing environments in the background to make correct control decisions for the UAV. 


We first tried with the MediaPipe~\cite{lugaresi2019mediapipe, sanchez2023lightweight} framework for basic hand detection coupled with a CNN for gesture classification. This CNN was designed with multiple convolutional layers and max-pooling for feature extraction, followed by dropout layers to mitigate overfitting. While the model achieved impressive test accuracy of $96.14\%$ during training, it resulted in significant reduction in validation accuracy. This model’s performance in real-time applications suffered, particularly in settings with variable background noise and greater detection distances. Precision, recall, and F1-scores were low, indicating potential misclassifications in real-world scenarios. Overall, the CNN model lacked adaptability to dynamic environments, and its inability to generalize well at longer distances limited its practical application.

The second approach we tried with improved upon the first by introducing segmentation-based detection, aiming to remove background interference. Here, a ResNet50 model served as the core classifier, using pre-trained ImageNet weights with fully connected layers removed. This architecture was then customized with additional layers—Flatten, Dropout, and Dense layers with six output units representing gesture classes. While segmentation increased accuracy initially, residual background noise persisted even after segmentation, resulting in misclassifications. Training accuracy progressively increased from $74.80\%$ to $97.60\%$, and validation accuracy rose to $99.89\%$. Despite these improvements, the ResNet50-based model struggled to generalize across varying backgrounds and distances, necessitating further exploration of techniques to reduce noise and overfitting.

We also tried with a third approach using a color-based segmentation with blue gloves to distinguish hand gestures against the background. This segmentation method aimed to isolate hand images effectively by assigning a distinct color to the hand. Despite some progress, this approach remained sensitive to backgrounds with blue shades, which introduced substantial noise and interfered with the model’s classification accuracy. Although color segmentation initially seemed effective, the results remained inconsistent, particularly across varying distances. The second and third approaches indicated that segmentation alone was insufficient to address background noise and maintain performance across extended distances.

The limitations of these initial approaches highlighted the need for a robust, background-agnostic solution capable of recognizing gestures consistently across distances. Landmark-based gesture recognition emerged as the most promising method, employing hand landmarks for gesture classification, which proved to be less susceptible to background noise and reliable over extended ranges.

\begin{figure}[!t]
    \centering
    \includegraphics[width=0.75\linewidth]{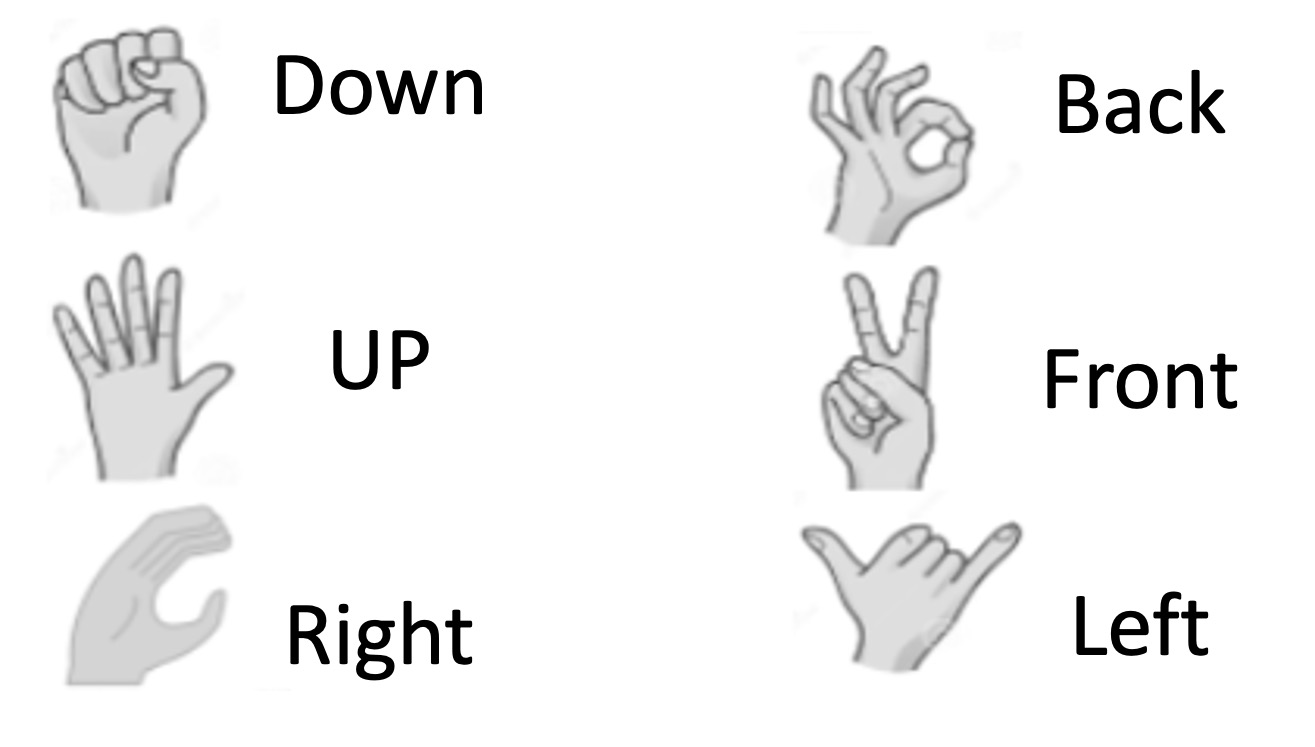}
    \caption{Mapping of different hand poses to UAV control commands}
    \label{fig:gesture_mapping}
    \vspace{-3mm}
\end{figure}

\vspace{-1mm}
\subsection{Landmark-Based Gesture Recognition}

To achieve accurate and robust hand gesture recognition for UAV control, we developed a custom landmark-based model. Our approach integrates MediaPipe’s hand landmark detection model with a custom-trained neural network classifier to recognize hand gestures effectively in real time. This combined method enables precise and consistent recognition regardless of environmental background or lighting variability.

Our custom landmark model is trained on a locally collected dataset specifically tailored for six hand gesture classes, each containing $1,500$ images. The collected dataset represents various hand poses used to communicate different commands to the UAV as shown in the figure~\ref{fig:gesture_mapping}, and the images were captured under diverse lighting conditions and from varying distances, enhancing model generalizability.

\subsection*{\textbf{Neural Network Architecture}}

\begin{figure}[!t]
    \centering
    \includegraphics[width=0.98\linewidth]{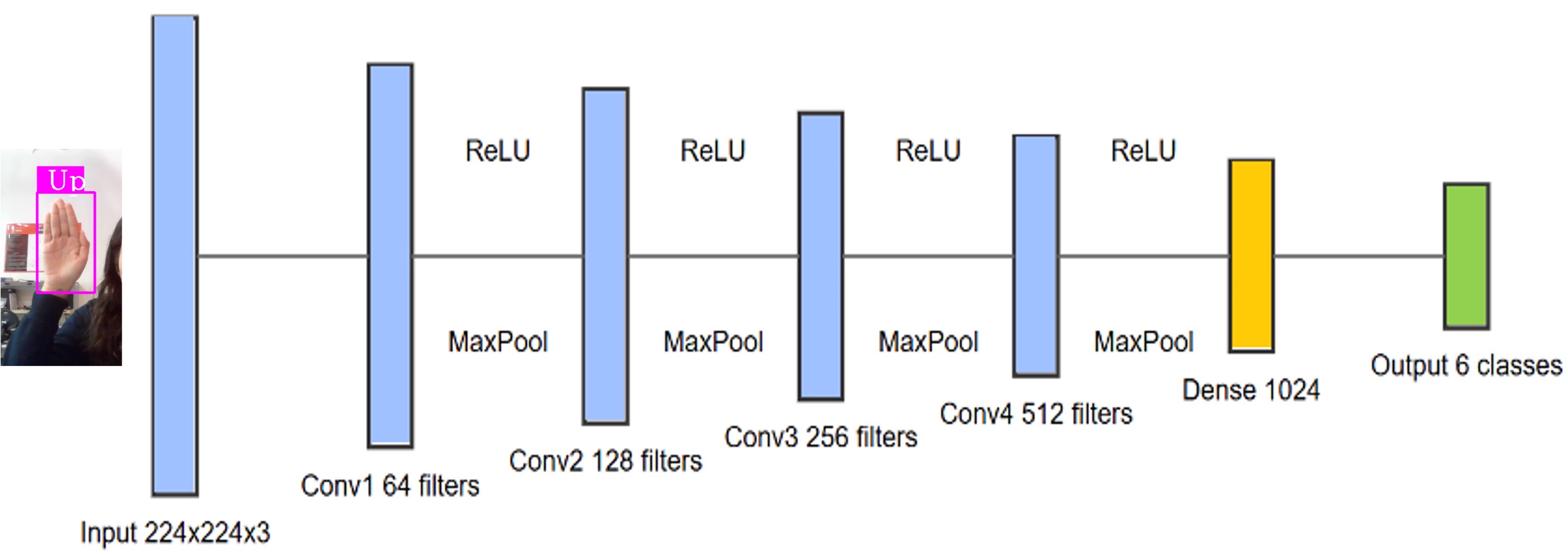}
    \caption{Neural Network Architecture for Hand Gesture Classification}
    \label{fig:network_architecture}
\end{figure}

The neural network classifier comprises four convolutional layers with ReLU activation functions for feature extraction, followed by two fully connected dense layers for classification. Figure \ref{fig:network_architecture} illustrates the architecture where the first convolutional layer extracts low-level features such as edges, while the deeper convolutional layers capture complex patterns associated with each gesture. Each convolutional layer is followed by a max-pooling layer to reduce dimensionality, thereby minimizing computational load and latency during real-time operation.


The input image $\mathbf{X}$ of dimensions $(224, 224, 3)$ undergoes convolution operations to generate feature maps $\mathbf{F}$, as defined by:
\begin{equation}
    \mathbf{F}_{l} = \sigma \left( \mathbf{W}_{l} * \mathbf{X}_{l-1} + \mathbf{b}_{l} \right)
\end{equation}

where $\mathbf{W}_{l}$ and $\mathbf{b}_{l}$ are weights and biases for layer $l$, $\sigma$ denotes the ReLU activation, and $*$ represents convolution. Max-pooling, applied after each convolution, reduces spatial dimensions:
\begin{equation}
    \mathbf{P}_{l} = \text{max\_pool}\left( \mathbf{F}_{l} \right)
\end{equation}

After feature extraction, the flattened final layer $\mathbf{P}_{\text{final}}$ is passed to fully connected layers. The output layer, using softmax, yields probabilities for each gesture class:
\begin{equation}
    \mathbf{y}_{\text{pred}} = \text{softmax} \left( \mathbf{W}_{\text{fc}} \cdot \mathbf{P}_{\text{final}} + \mathbf{b}_{\text{fc}} \right)
\end{equation}

where $\mathbf{W}_{\text{fc}}$ and $\mathbf{b}_{\text{fc}}$ represent weights and biases of the output layer. This configuration supports accurate classification based on the highest probability.

\subsection*{\textbf{Hyperparameter Tuning}}

To improve performance and latency, we conducted extensive hyperparameter tuning, optimizing learning rate, batch size, and dropout rate. Testing a range of learning rates from $0.0001$ to $0.01$, we selected $0.001$, which led to stable convergence. With a batch size of $32$, memory efficiency and convergence speed were balanced, while a dropout rate of $0.4$ mitigated overfitting in dense layers. These refinements improved generalization and enabled real-time performance.

\subsection*{\textbf{Training Process}}

Training of the model utilized the Adam optimizer and categorical cross-entropy loss, minimizing the loss over $20$ epochs. Early stopping prevented overfitting. After training, the model achieved a testing accuracy of $96.14\%$ on our dataset, confirming its effectiveness for real-time UAV control.

\subsection*{\textbf{Extended Distance with YOLO Integration}}
One of the major challenges in the gesture recognition based UAV control is that if the UAV goes further and further away from the user, then the images captured by the drone camera will be more noise or blurry, causing inaccuracy or failures in gesture recognition. Most of the existing models that we used earlier could only detect the gestures correctly up to a few meters and after finetuning up to $5$ meters.
To further enhance the recognition range, we experimented with the YOLOv4 model~\cite{bochkovskiy2020yolov4, aziz2023yolo} in place of MediaPipe. This integration allowed for accurate gesture recognition from $1$ to $10$ meters, surpassing the original $1–5$ meter range. The YOLOv4 framework detects hands and extracts regions of interest, which are then fed into our trained landmark model. This modification, enabled by YOLO’s object detection capabilities, significantly improves recognition distance, thus extending UAV operability in wider spaces.

\vspace{2mm}
\section{Edge-Assisted Distributed Computing}
Running a gesture detection model in real-time on compact drones, such as the DJI Tello, or using microcomputers like the Raspberry Pi or other embedded computing boards introduces challenges due to hardware limitations~\cite{spicer2023performance}. The onboard processing power of these devices often struggles with the computational demands of high-frequency neural network inference, especially for tasks like hand landmark detection and classification. Latency becomes a critical issue as processing delays directly affect responsiveness, compromising the precision needed for effective UAV control. 
In our trials, running the hand landmark model directly on DJI Tello or a Raspberry Pi led to significant latency, with up to $500$\,ms delays in command response times. This lag becomes problematic in real-time UAV control, where low-latency gesture recognition is essential to ensure safe, responsive flight. 

Edge computing~\cite{callegaro2019information, callegaro2020dynamic, callegaro2019measurement1} presents a solution to address these latency issues by offloading intensive computations to nearby computing units that are typically one-hop wireless distance away from the UAVs. A laptop or PC can work as an edge-computing unit, reducing the processing load on the drone itself and enhancing real-time performance. In this approach, the drone captures the video feed and streams it over wireless network, e.g., WiFi to an edge server (in this case, a laptop or PC) for processing. Equipped with optimized models and greater computational resources, the edge server handles gesture detection much faster and sends back the control commands to the drone. This edge-assisted distributed computing allows for faster inference times while keeping the drone agile, enabling the use of more complex models without overloading the onboard processor. 

\begin{figure}[!t]
    \centering
    \includegraphics[width=0.80\linewidth]{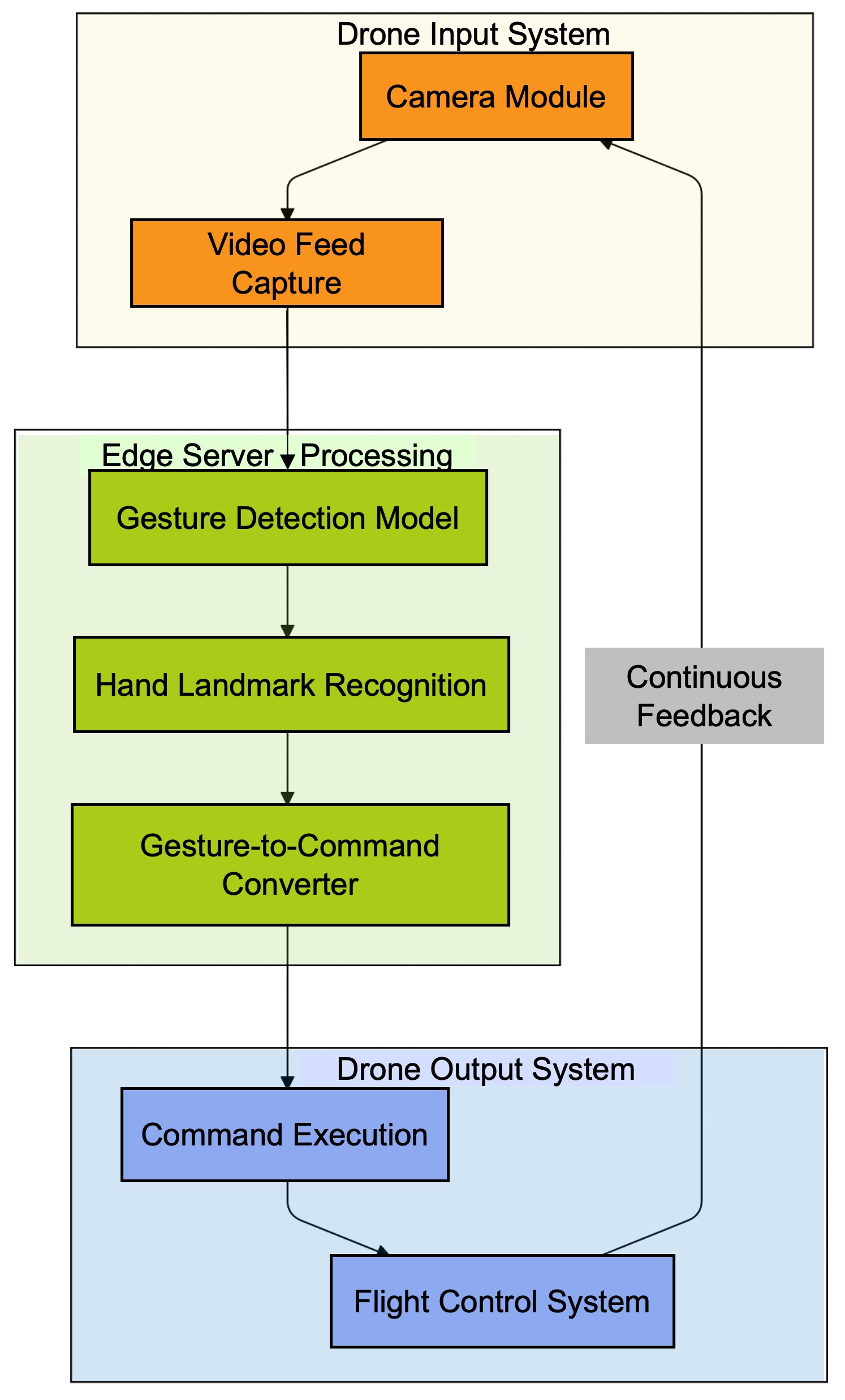} 
    \caption{End-to-end System setup for Edge-assisted Gesture-recognition based UAV Control}
    \label{fig:edge_computing}
\end{figure}

Figure~\ref{fig:edge_computing} illustrates the end-to-end architecture showing the data flow and processing steps in this edge-computing setup.
This architecture distributes the processing steps between the drone and the edge server, where the server handles the deep learning based gesture detection and landmark recognition, then transmits only the control commands to the drone. By offloading the resource-intensive tasks to the edge computing node, this setup ensures an efficient, low-latency gesture recognition and hence a real-time interaction between the user and the UAV with minimal latency.

\subsection{Communication Protocol}
To enable the edge computing in the framework, The UAV needs to be connected with the edge computing server over wireless network.
We implement the system utilizing DJI Tello drone's built-in WiFi access point operating at $2.4$\,GHz, with the {\it djitellopy} Python library facilitating the communication. The protocol implementation includes the following:

\begin{itemize}
    \item \textbf{Video Streaming}: UDP port 11111 for receiving 720p video feed (15 FPS).
    \item \textbf{Command Communication}: UDP port 8889 for sending control commands.
    \item \textbf{State Information}: UDP port 8890 for receiving drone telemetry data.
    \item \textbf{Connection Monitoring}: Automatic detection of connection loss with 10ms timeout.
\end{itemize}

The system maintains reliable communication through maximum command rate of $10$\,Hz to prevent buffer overflow. 
It also employs automatic video frame dropping during high latency which can be caused by variation in the wireless channel conditions due to environmental dynamics. 
The system implements a command queue management for smooth execution and battery level monitoring with automatic landing at $15\%$ remaining energy.

\subsection{Enhanced Robustness with Failsafe Mechanisms}
The system implements multiple failsafe features to ensure the operation is safe if the system fails at one or more points. First, when gestures are unclear,i.e., the detection confidence is low, the UAV triggers automatic hover mode. If the connection with the edge server is lost for a time threshold, it will call return-to-home functionality. For safe operation, one can limit the maximum distance by incorporating geo-fencing. And at any point if the battery level goes below a threshold, it will land to ensure safety.


\vspace{2mm}
\section{Performance Evaluation}

\subsection{Experiment Setup}

The experiments were conducted in both simulated and real-world environments to evaluate the hand gesture recognition system for UAV control across varying distances and environmental conditions. We used both AirSim for simulation and a DJI Tello drone for real-world testing. The experimental setup aimed to simulate UAV operations in dynamic environments, where precise hand gesture recognition could be used for remote control and command execution.

\begin{figure}[!t]
    \centering
    \vspace{2mm}\includegraphics[width=0.5\textwidth]{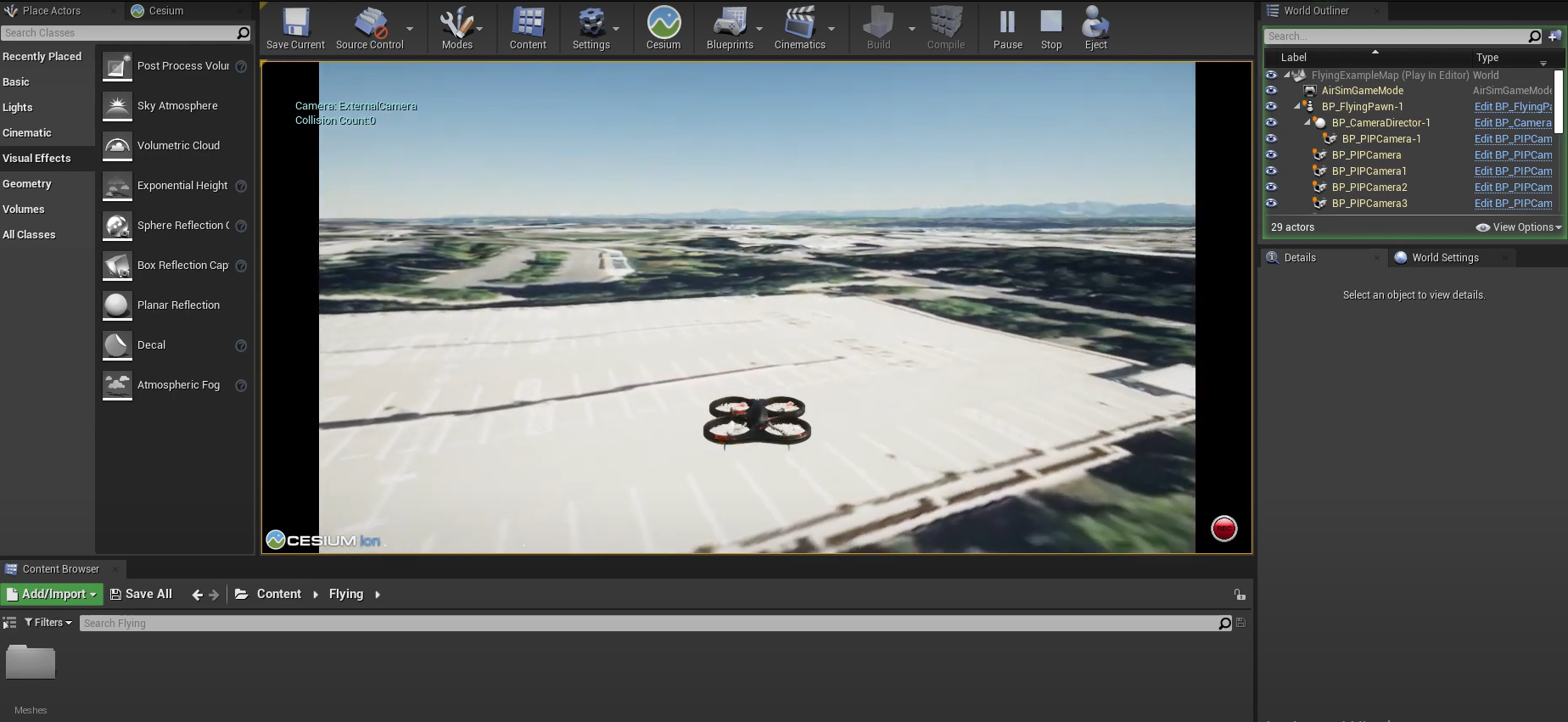}
    \caption{Experiment setup for gesture-based UAV control using AirSim}
    \label{fig:experiment_setup}
    \vspace{-3mm}
\end{figure}

Table~\ref{table:experiment} summarizes the parameters setup used in the experiments.
In the AirSim simulation environment, a virtual network radius of $400$\,m was established, with UAV movement paths designed for automated gesture-based command reception as shown in figure~\ref{fig:experiment_setup}. In real-world testing, the DJI Tello drone provided real-time video feed in 720p resolution, transmitting video input to an edge server (Laptop/PC) for processing. The video feed was processed to detect and classify gestures in real time, enabling dynamic UAV control.
Throughout all tests, we used a landmark-based hand detection model augmented by YOLOv4 for detecting gestures over extended ranges. Experiments were conducted across a range of distances from $1$ to $10$ meters. The system was implemented using Python, leveraging OpenCV and TensorFlow libraries for video processing, with MediaPipe employed for hand landmark detection.

\begin{table}[!t]
    \centering
    \vspace{2mm}
    \begin{tabular}{|l|c|c|}
        \hline
        \textbf{Parameters} & \textbf{Symbols} & \textbf{Values} \\
        \hline
        Edge Server & - & Laptop/PC \\
        Frame Resolution & $R$ & 1280 x 720 pixels (720p) \\
        Model Input Image Size & $S$ & 224 x 224 pixels \\
        Latency Requirement & $\tau$ & < 30 ms \\
        Max Detection Range & $d_{\text{max}}$ & 10 meters \\
        Drone Operating Altitude & $h$ & 1-3 meters \\
        Reliability (Detection Accuracy) & $\chi$ & 96.14\% \\
        Classification Model & - & YOLOv4 + Landmark \\
        \hline
    \end{tabular}
    \caption{Specifications of the Gesture Detection Experiment}
    \label{table:experiment}
\end{table}

In this setup, the Tello drone captures real-time video at 720p resolution, transmitting it to the edge server for processing. The edge server performs landmark-based hand detection and gesture classification using the YOLOv4-enhanced model. Control commands are then transmitted back from the edge server to the drone based on the recognized gestures, allowing for responsive and dynamic UAV control even at extended distances.

\section{Results}
Our experimental evaluation focuses on three key aspects: recognition accuracy, system latency, and trajectory tracking performance.

\subsection{Gesture Recognition Performance}
\begin{figure}[!t]
    \centering
    \includegraphics[width=0.99\linewidth]{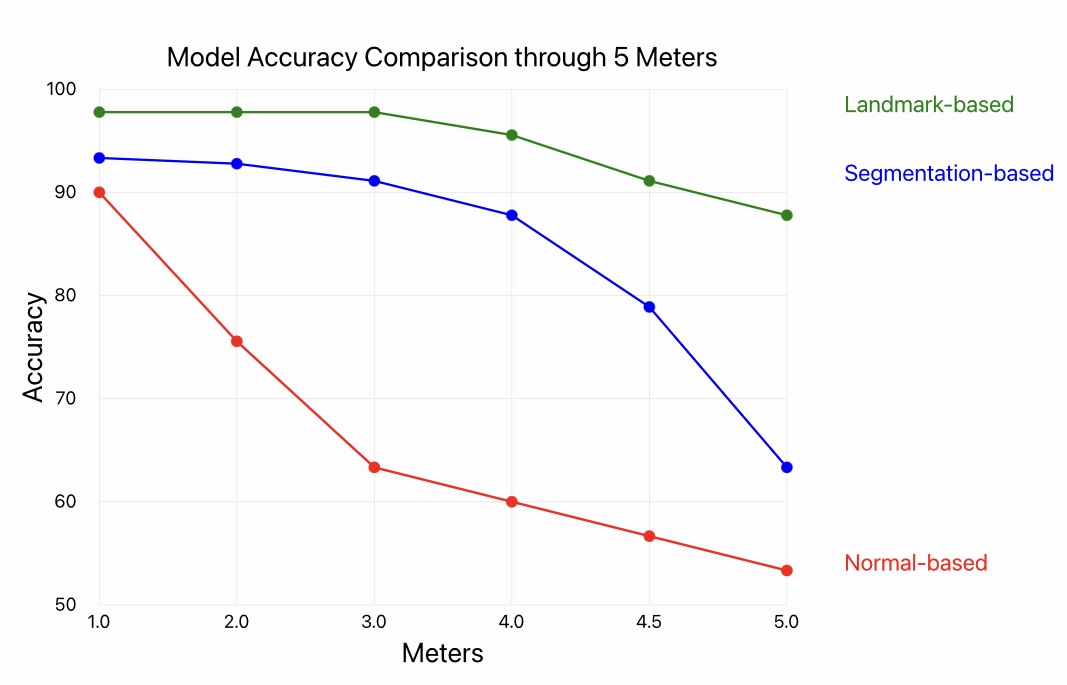}
    \caption{Accuracy comparison}
    \label{fig:result_acc}
\end{figure}

The landmark-based approach demonstrated superior performance compared to traditional methods, as shown in Figure~\ref{fig:result_acc}. It can be seen from the plot that the landmark based gesture recognition significantly outperforms the normal CNN-based model and also the CNN with segmentation.
We got an overall Accuracy of $96.14\%$ for landmark-based detection across all gesture classes. It also shows that as we vary the distance from $1m$ to $5m$, the landmark-based method although slightly degrades in accuracy due to added noises, it maintains almost $90\%$ accuracy up to $5m$ distance. In contrary, the accuracy of the other two models sharply decreases with distance after $1$ or $2$ meters. 

We also measure the accuracy of the landmark-based gesture recognition model under varying lighting conditions and observed less than $5\%$ degradation in performance.
Additionally, the landmark-based model incurs a very low false positive rate (below $2\%$) compared to other methods. As we mentioned that our system handles false negative by hovering, but the false positive can highly impact as it can create error in trajectory and/or mission failures.

\subsection{Latency Performance}
The end-to-end closed-loop latency comprises the three major latency components -- the video streaming delay from the UAV to the edge server, the edge processing delay, the network delay to send the command from the edge server to the UAV. We measure these individual components along with the end-to-end latency as well.

\vspace{1mm}
\noindent
{\bf Video Stream Latency:} The video data transmission speed depends on the resolution of the captured images and the network bandwidth. Since we used WiFi network and $720$p resolution, we get around $80$\,ms to $120$\,ms latency for video streaming from the drone camera to the edge server.

\begin{figure}[!t]
    \centering
    \includegraphics[width=0.5\textwidth]{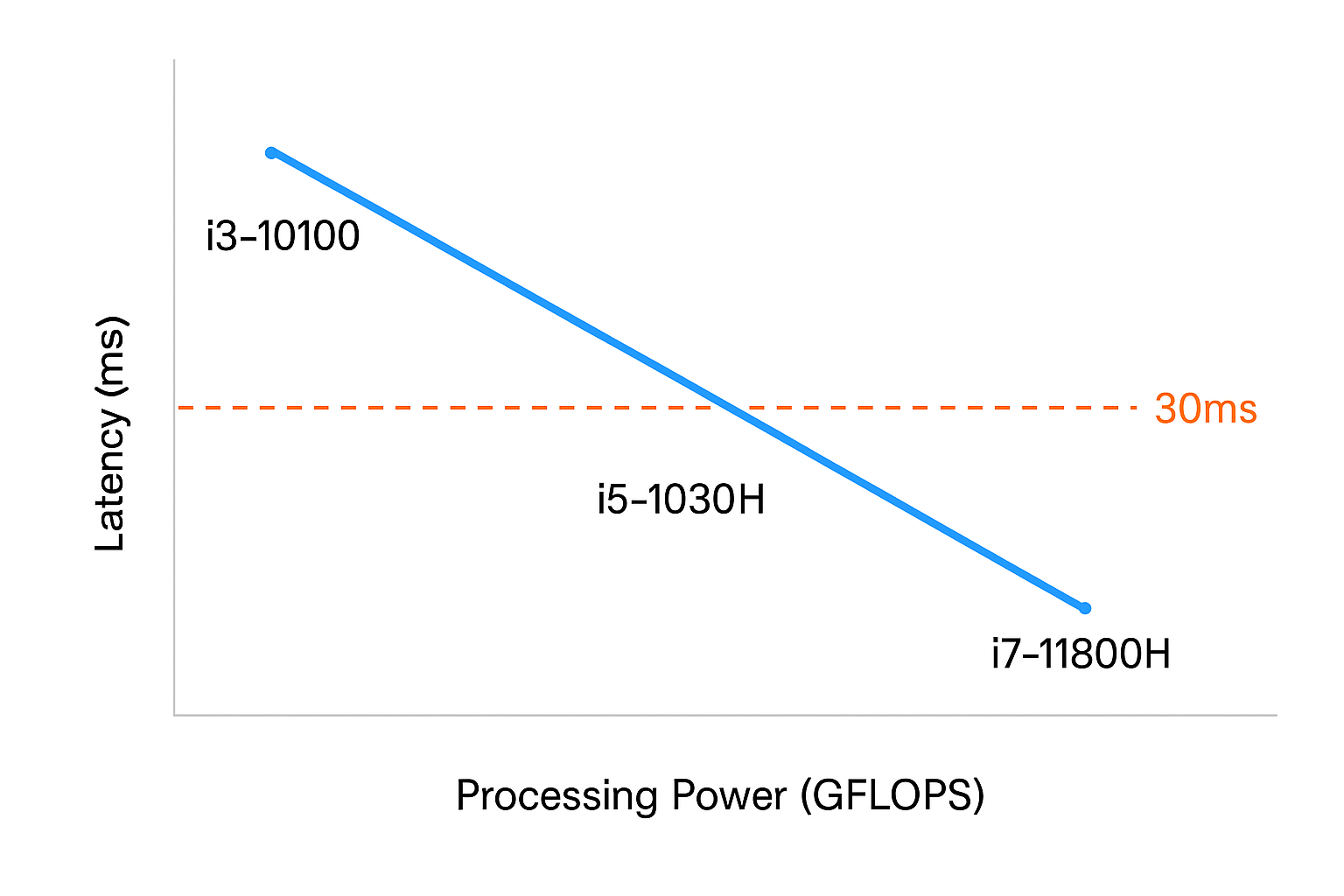}
    \caption{System Latency vs Edge Computing Capacity}
    \label{fig:results_latency}
\end{figure}

\vspace{1mm}
\noindent
{\bf Edge Processing Latency:} 
Figure~\ref{fig:results_latency} illustrates the processing latency across different edge computing configurations. We experimented with different CPU hardware and run our landmark-based model and the conversion to control commands. We got 22ms average latency (30 FPS) on i7-11800H processor, 28ms average latency (25 FPS) on i5-10300H processor, and 35ms average latency (20 FPS) on i3-10100 processor respectively.

\vspace{1mm}
\noindent
{\bf Command Transmission Delay:}
Since the converted commands from the detected gesture contains a very small amount of data which is transmitted over the wireless network from the edge server to the UAV, it takes a very small amount of time. We measure an 
average of  $12$\,ms round-trip time for command transmission.

\vspace{1mm}
\noindent
{\bf End-to-End Response Latency}: 
When we measure the sensing to control end-to-end latency, we get around $150$\,ms from video transmission to gesture detection to drone movement. This latency is low enough to provide a real-time experiment as the user creating the gestures is a human in the loop and the this speed is typically faster than human response speed.

\vspace{2mm}
\subsection{Flight Control Performance}
\begin{figure}[!t]
    \centering
    \includegraphics[width=0.5\textwidth]{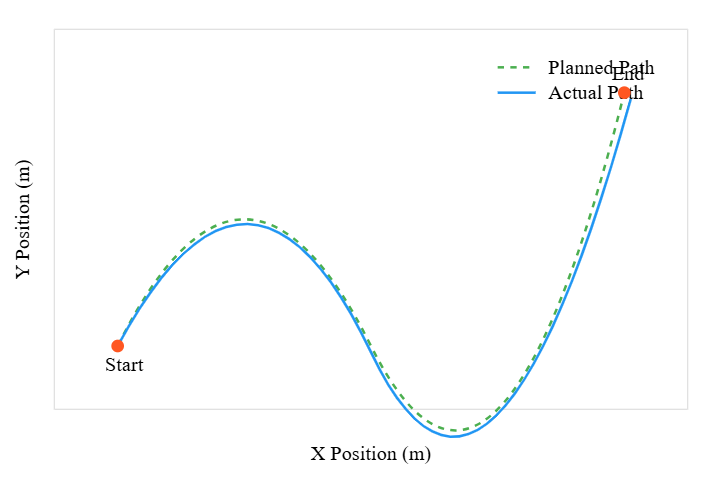}
    \caption{Drone Trajectory Tracking}
    \label{fig:results_trajectory}
\end{figure}

We also measure the  deviation of the UAV from the planned trajectory. 
Figure~\ref{fig:results_trajectory} demonstrates the system's control precision in terms of position accuracy, path following accuracy and stability.

\vspace{1mm}
\noindent
{\bf Position Accuracy}:
We measure the position accuracy of the UAV in different operating modes. In the hover mode, we get $\pm 15$\,cm deviation, whereas in the forward flight, it gets about $\pm 25$\,cm lateral deviation. During the turn maneuvers, the heading accuracy can be deviated by 
$\pm 5$°. 

\vspace{1mm}
\noindent
{\bf Path Following}:
Due to the errors in gesture recognition, wrong commands can get triggered and as the commands play out through a queue, it can accumulate errors in terms of deviation in trajectory from the planned path. By using our edge-assisted landmark-based gesture recognition model, we obtain a  $92\%$ trajectory accuracy for planned paths. We also measure the average deviation of the UAV from the planned route and it is also very small ($18$\,cm) as shown in figure~\ref{fig:results_trajectory}
As far as the successful completion of the commands is concerned, it gets successful in $96\%$ of the time.

\vspace{1mm}
\noindent
{\bf Performance with Environmental Dynamics}:
Finally, we test our performance in dynamic environmental conditions. We get stable operation in winds up to $8.3$\,m/s speed and 
reliable gesture recognition in $50-1000$ lux lighting. As far as the control range is concerned, our landmark-based gesture recognition model can provide an effective control range up to $10$ meters which is more than any existing computer vision based gesture recognition method.

These results demonstrate that our edge-assisted gesture recognition system achieves both the accuracy and responsiveness required for practical UAV control applications. The landmark-based approach, combined with edge computing, provides robust performance across varying environmental conditions while maintaining latency within acceptable limits for real-time control.

\section{Conclusions}

This study presents a novel approach for real-time hand gesture recognition based UAV control through a landmark-based model combined with the YOLOv4 object detection framework. By leveraging edge computing, we mitigated the latency issues typically associated with real-time processing of these complex tasks on UAVs, achieving a reliable gesture recognition within a $30$\,ms latency threshold. The system was tested across a range of environments and showed robustness in varying conditions and distances, confirming its effectiveness for responsive UAV control. These findings indicate that edge-assisted landmark-based gesture recognition offers a practical and efficient solution for remote UAV control in real-world scenarios.

\section*{Acknowledgment}
This work is partially supported by the US National Science Foundation (NSF EPSCoR \#1849213).

\balance

\bibliographystyle{IEEEtran}
\bibliography{references}

\begin{thebibliography}{10}
\providecommand{\url}[1]{#1}
\csname url@samestyle\endcsname
\providecommand{\newblock}{\relax}
\providecommand{\bibinfo}[2]{#2}
\providecommand{\BIBentrySTDinterwordspacing}{\spaceskip=0pt\relax}
\providecommand{\BIBentryALTinterwordstretchfactor}{4}
\providecommand{\BIBentryALTinterwordspacing}{\spaceskip=\fontdimen2\font plus
\BIBentryALTinterwordstretchfactor\fontdimen3\font minus \fontdimen4\font\relax}
\providecommand{\BIBforeignlanguage}[2]{{%
\expandafter\ifx\csname l@#1\endcsname\relax
\typeout{** WARNING: IEEEtran.bst: No hyphenation pattern has been}%
\typeout{** loaded for the language `#1'. Using the pattern for}%
\typeout{** the default language instead.}%
\else
\language=\csname l@#1\endcsname
\fi
#2}}
\providecommand{\BIBdecl}{\relax}
\BIBdecl

\bibitem{muchiri2022review}
G.~Muchiri and S.~Kimathi, ``A review of applications and potential applications of uav,'' in \emph{Proceedings of the Sustainable Research and Innovation Conference}, 2022, pp. 280--283.

\bibitem{daud2022applications}
S.~M. S.~M. Daud, M.~Y. P.~M. Yusof, C.~C. Heo, L.~S. Khoo, M.~K.~C. Singh, M.~S. Mahmood, and H.~Nawawi, ``Applications of drone in disaster management: A scoping review,'' \emph{Science \& Justice}, vol.~62, no.~1, pp. 30--42, 2022.

\bibitem{elmokadem2021towards}
T.~Elmokadem and A.~V. Savkin, ``Towards fully autonomous uavs: A survey,'' \emph{Sensors}, vol.~21, no.~18, p. 6223, 2021.

\bibitem{stegagno2014semi}
P.~Stegagno, M.~Basile, H.~H. B{\"u}lthoff, and A.~Franchi, ``A semi-autonomous uav platform for indoor remote operation with visual and haptic feedback,'' in \emph{2014 IEEE international conference on robotics and automation (ICRA)}.\hskip 1em plus 0.5em minus 0.4em\relax IEEE, 2014, pp. 3862--3869.

\bibitem{patrona2021overview}
F.~Patrona, I.~Mademlis, and I.~Pitas, ``An overview of hand gesture languages for autonomous uav handling,'' \emph{2021 Aerial Robotic Systems Physically Interacting with the Environment (AIRPHARO)}, pp. 1--7, 2021.

\bibitem{yu2022end}
C.~Yu, S.~Fan, Y.~Liu, and Y.~Shu, ``End-side gesture recognition method for uav control,'' \emph{IEEE Sensors Journal}, vol.~22, no.~24, pp. 24\,526--24\,540, 2022.

\bibitem{yoo2022motion}
M.~Yoo, Y.~Na, H.~Song, G.~Kim, J.~Yun, S.~Kim, C.~Moon, and K.~Jo, ``Motion estimation and hand gesture recognition-based human--uav interaction approach in real time,'' \emph{Sensors}, vol.~22, no.~7, p. 2513, 2022.

\bibitem{jiang2019toward}
C.~Jiang, X.~Cheng, H.~Gao, X.~Zhou, and J.~Wan, ``Toward computation offloading in edge computing: A survey,'' \emph{IEEE Access}, vol.~7, pp. 131\,543--131\,558, 2019.

\bibitem{shah2018airsim}
S.~Shah, D.~Dey, C.~Lovett, and A.~Kapoor, ``Airsim: High-fidelity visual and physical simulation for autonomous vehicles,'' in \emph{Field and Service Robotics: Results of the 11th International Conference}.\hskip 1em plus 0.5em minus 0.4em\relax Springer, 2018, pp. 621--635.

\bibitem{hu2020deep}
B.~Hu and J.~Wang, ``Captainglove: Capacitive and inertial fusion-based glove for real-time on edge hand gesture recognition for drone control,'' \emph{International Journal of Automation and Computing}, vol.~17, pp. 17--29, 2020.

\bibitem{samotyy2024gesture}
V.~Samotyy, N.~Kiselov, U.~Dzelendzyak, and O.~Shpak, ``Gesture recognition based on deep learning for quadcopters flight control,'' \emph{International Journal of Computing}, vol.~17, pp. 17--29, 2024.

\bibitem{bello2023captainglove}
H.~Bello, S.~Suh, D.~Geißler, L.~Ray, and P.~Zhou, B.~andLukowicz, ``Gesture recognition based on deep learning for quadcopters flight control,'' \emph{preprint arXiv:2306.04319}, 2023.

\bibitem{perera2018uav}
A.~G. Perera, Y.~W. Law, and J.~Chahl, ``Uav-gesture: A dataset for uav control and gesture recognition,'' \emph{Proceedings of the European Conference on Computer Vision (ECCV) Workshops}, pp. 227--243, 2018.

\bibitem{lee2023wearable}
J.~Lee and K.~H. Yu, ``Wearable drone controller: Machine learning-based hand gesture recognition and vibrotactile feedback,'' \emph{Sensors}, vol.~5, 2023.

\bibitem{lugaresi2019mediapipe}
C.~Lugaresi, J.~Tang, H.~Nash, C.~McClanahan, E.~Uboweja, M.~Hays, F.~Zhang, C.-L. Chang, M.~Yong, J.~Lee \emph{et~al.}, ``Mediapipe: A framework for perceiving and processing reality,'' in \emph{Third workshop on computer vision for AR/VR at IEEE computer vision and pattern recognition (CVPR)}, vol. 2019, 2019.

\bibitem{sanchez2023lightweight}
G.~S{\'a}nchez-Brizuela, A.~Cisnal, E.~de~la Fuente-L{\'o}pez, J.-C. Fraile, and J.~P{\'e}rez-Turiel, ``Lightweight real-time hand segmentation leveraging mediapipe landmark detection,'' \emph{Virtual Reality}, vol.~27, no.~4, pp. 3125--3132, 2023.

\bibitem{bochkovskiy2020yolov4}
A.~Bochkovskiy, C.-Y. Wang, and H.-Y.~M. Liao, ``Yolov4: Optimal speed and accuracy of object detection,'' \emph{arXiv preprint arXiv:2004.10934}, 2020.

\bibitem{aziz2023yolo}
A.~B. Aziz, N.~Basnin, M.~Farshid, M.~Akhter, T.~Mahmud, K.~Andersson, M.~S. Hossain, and M.~S. Kaiser, ``Yolo-v4 based detection of varied hand gestures in heterogeneous settings,'' in \emph{International Conference on Applied Intelligence and Informatics}.\hskip 1em plus 0.5em minus 0.4em\relax Springer, 2023, pp. 325--338.

\bibitem{spicer2023performance}
E.~Spicer and S.~Baidya, ``Performance tradeoff in dnn-based coexisting applications in resource-constrained cyber-physical systems,'' in \emph{2023 IEEE International Conference on Smart Computing (SMARTCOMP)}.\hskip 1em plus 0.5em minus 0.4em\relax IEEE, 2023, pp. 219--221.

\bibitem{callegaro2019information}
D.~Callegaro, S.~Baidya, G.~S. Ramachandran, B.~Krishnamachari, and M.~Levorato, ``Information autonomy: Self-adaptive information management for edge-assisted autonomous uav systems,'' in \emph{MILCOM 2019-2019 IEEE Military Communications Conference (MILCOM)}.\hskip 1em plus 0.5em minus 0.4em\relax IEEE, 2019, pp. 40--45.

\bibitem{callegaro2020dynamic}
D.~Callegaro, S.~Baidya, and M.~Levorato, ``Dynamic distributed computing for infrastructure-assisted autonomous uavs,'' in \emph{ICC 2020-2020 IEEE International Conference on Communications (ICC)}.\hskip 1em plus 0.5em minus 0.4em\relax IEEE, 2020, pp. 1--6.

\bibitem{callegaro2019measurement1}
------, ``A measurement study on edge computing for autonomous uavs,'' in \emph{Proceedings of the ACM SIGCOMM 2019 Workshop on Mobile AirGround Edge Computing, Systems, Networks, and Applications}.\hskip 1em plus 0.5em minus 0.4em\relax IEEE, 2019, pp. 29--35.

\end{thebibliography}


\begin{thebibliography}{99}

\bibitem{hu2020deep}
B. Hu and J. Wang, "Deep Learning Based Hand Gesture Recognition and UAV Flight Controls," \textit{International Journal of Automation and Computing}, vol. 17, no. 1, pp. 17--29, 2020. [Online]. Available: \url{https://doi.org/10.1007/s11633-019-1194-7}

\bibitem{samotyy2024gesture}
V. Samotyy, N. Kiselov, U. Dzelendzyak, and O. Shpak, "Gesture Recognition based on Deep Learning for Quadcopters Flight Control," \textit{International Journal of Computing}, vol. 23, no. 4, 2024. [Online]. Available: \url{https://doi.org/10.47839/ijc.23.4.3757}

\bibitem{bello2023captainglove}
H. Bello, S. Suh, D. Geißler, L. Ray, B. Zhou, and P. Lukowicz, "CaptAinGlove: Capacitive and Inertial Fusion-Based Glove for Real-Time on Edge Hand Gesture Recognition for Drone Control," \textit{arXiv preprint arXiv:2306.04319}, 2023. [Online]. Available: \url{https://arxiv.org/abs/2306.04319}

\bibitem{perera2018uav}
A. G. Perera, Y. W. Law, and J. Chahl, "UAV-GESTURE: A Dataset for UAV Control and Gesture Recognition," in \textit{Proceedings of the European Conference on Computer Vision (ECCV) Workshops}, 2018, pp. 227--243. [Online]. Available: \url{https://openaccess.thecvf.com/content_ECCVW_2018/papers/11130/Perera_UAV-GESTURE_A_Dataset_for_UAV_Control_and_Gesture_Recognition_ECCVW_2018_paper.pdf}

\bibitem{lee2023wearable}
J.-W. Lee and K.-H. Yu, "Wearable Drone Controller: Machine Learning-Based Hand Gesture Recognition and Vibrotactile Feedback," \textit{Sensors}, vol. 23, no. 5, p. 2575, 2023. [Online]. Available: \url{https://www.mdpi.com/1424-8220/23/5/2575}

\end{thebibliography}

\end{document}